\pdfoutput=1

\documentclass[11pt]{article}

\usepackage[]{acl}

\usepackage{times}
\usepackage{latexsym}
\usepackage{graphicx}

\usepackage[T1]{fontenc}

\usepackage[utf8]{inputenc}

\usepackage{microtype}

\usepackage{inconsolata}

\usepackage{booktabs}
\usepackage{multirow}
\usepackage{pgfplots}
\usepackage{url}

\definecolor{color1}{RGB}{159,168,218}
\definecolor{color2}{RGB}{173,201,198}
\definecolor{color3}{RGB}{207,216,220}
\definecolor{color4}{RGB}{247,194,183}
\definecolor{color5}{RGB}{187,230,190}
\definecolor{color6}{RGB}{187,230,222}

\usepackage{amsmath}

\title{KG-Rank: Enhancing Large Language Models for Medical QA \\
with Knowledge Graphs and Ranking Techniques   
}

\author{Rui Yang$^{1*}$, Haoran Liu$^2$, Edison Marrese-Taylor $^3$, Qingcheng Zeng $^4$, Yu He Ke$^5$, Wanxin Li$^6$, \\ \textbf{Lechao Cheng$^7$, Qingyu Chen$^{8,9}$, James Caverlee$^2$, Yutaka Matsuo$^3$, Irene Li$^{3*}$} \\
\\
$^1$Duke-NUS Medical School, $^2$Texas A\&M University, 
$^3$The University of Tokyo, \\
$^4$Northwestern University, $^5$Singapore General Hospital, $^6$Zhejiang University,\\
$^7$Zhejiang Lab, $^8$Yale University, 
$^9$National Institutes of Health \\
yang.rui@duke-nus.edu.sg, ireneli@ds.itc.u-tokyo.ac.jp
}
\pgfplotsset{compat=1.18}
\begin{document}
\maketitle
\begin{abstract}
Large language models (LLMs) have demonstrated impressive generative capabilities with the potential to innovate in medicine. However, the application of LLMs in real clinical settings remains challenging due to the lack of factual consistency in the generated content. In this work, we develop an augmented LLM framework, KG-Rank, which leverages a medical knowledge graph (KG) along with ranking and re-ranking techniques, to improve the factuality of long-form question answering (QA) in the medical domain. Specifically, when receiving a question, KG-Rank automatically identifies medical entities within the question and retrieves the related triples from the medical KG to gather factual information. Subsequently, KG-Rank innovatively applies multiple ranking techniques to refine the ordering of these triples, providing more relevant and precise information for LLM inference. To the best of our knowledge, KG-Rank is the first application of KG combined with ranking models in medical QA specifically for generating long answers. Evaluation on four selected medical QA datasets demonstrates that KG-Rank achieves an improvement of over 18\% in  ROUGE-L score. Additionally, we extend KG-Rank to open domains, including law, business, music, and history, where it realizes a 14\% improvement in ROUGE-L score, indicating the effectiveness and great potential of KG-Rank.
\end{abstract}

\section{Introduction}

Large language models (LLMs), such as GPT-4 \cite{openai2023gpt4} and LLaMa2 \cite{touvron2023llama}, have demonstrated powerful generative capabilities \cite{gao2023large, yang2024leveraging}. Despite their considerable potential in various domains, including medicine \cite{li2022neural, yang2023ascle, ke2024enhancing, yang2024retrieval}, their limited training on medical data raises concerns about the consistency of the generated content with established medical facts \cite{yang2023large, bi2024decodingcontrastingknowledgeenhancing}. 


To address this challenge without additional computational cost, previous research, such as Almanac \cite{hiesinger2023almanac} and ChatENT \cite{long2023chatent}, leverages external medical knowledge to enhance the accuracy and reliability of LLM-generated content. However, merely retrieving external knowledge risks introducing irrelevant or unreliable information \cite{yang2024retrieval}, which can compromise the effectiveness of LLMs, and raise issues of credibility, data consistency, privacy, security, and legality. While previous studies have emphasized the advantages of utilizing external knowledge, they have overlooked a crucial question: \textit{How to better integrate external knowledge?}

In this work, we propose \textbf{KG-Rank}, an augmented framework that integrates a structured medical knowledge graph (KG) with ranking techniques into LLMs to achieve more accurate and reliable long-form medical question-answering (QA). We first retrieve one-hop relations of related medical entities from the medical KG (Unified Medical Language System (UMLS)) \cite{bodenreider2004unified}. To retain relevant information from the KG, we then propose to apply ranking and re-ranking methods to optimize the ordering of triplets.

Specifically, we introduce three ranking techniques to improve the integration of LLM with KG by filtering irrelevant data, highlighting key information, and ensuring diversity. These techniques also streamline the process by reducing the number of triplets required for LLM inference. Additionally, we apply re-ranking models to reassess and emphasize the most relevant triplets, enhancing the factuality of KG-Rank in the long-form medical QA task.

To summarize, our contributions are: (1) We propose KG-Rank, a KG-augmented LLM framework for the medical QA task. To the best of our knowledge, this is the first application of KG combined with ranking techniques to enhance LLMs for medical QA with long answers. (2) We incorporate different ranking and re-ranking techniques to eliminate noise and redundancy in the KG-retrieval stage. (3) We validate the effectiveness of KG-Rank on both medical and various open-domain QA tasks. All the data and code can be found at \url{https://github.com/YangRui525/KG-Rank}.

\section{Methodology}
\begin{figure}[h]
  \centering
  \includegraphics[width=0.507\textwidth]{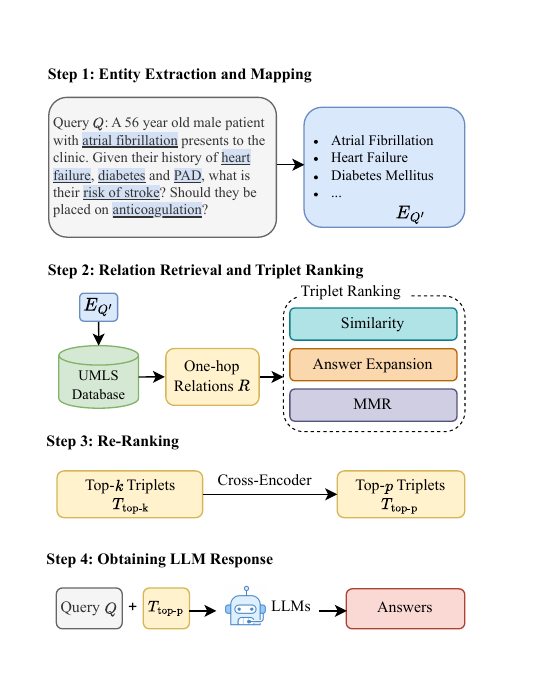} 
  \caption{An illustration of KG-Rank Framework. }
  \label{fig:framework}
  \vspace{-5mm}
\end{figure}

As shown in Fig.~\ref{fig:framework}, we introduce the KG-Rank (Knowledge Graph  -Rank) framework for the long-form medical QA task. 

\subsection{External Knowledge Graph}\label{sec: med_kg}
We define the external KG as \(G = (V, E)\), where \(V\) represents the set of entities and \(E\) represents the set of structural relations. For the medical QA task, we choose UMLS as the primary medical KG. UMLS is a comprehensive repository of health and biomedical vocabularies, designed to promote information standardization and interoperability. The core component of UMLS, the Metathesaurus, contains over 3.8 million concepts and more than 78 million relations, and supports 25 languages, providing extensive medical knowledge coverage to enhance LLMs. In UMLS, knowledge is represented in the form of triples, which consist of two medical concepts and the relation between them. For example, in the triple \textit{(Myopia, clinically\_associated\_with, HYPERGLYCEMIA)}, "Myopia" and "HYPERGLYCEMIA" are medical concepts, while "clinically\_associated\_with" is the relation between them.

\subsection{Entity Extraction and Mapping}\label{sec: entity_extract}

In the first step, we extract key entities and find mappings from the external KG. Specifically, for the given question \(Q\), we apply a Medical NER Prompt \(P_{\text{MedNER}}\) to identify related medical entities \(E_{Q}\), and then we map each entity \(e_i \in E_{Q}\) to the corresponding entity in the knowledge graph \(G\). The detailed prompt can be found in Appendix~\ref{med_prompt}.

\subsection{Relation Retrieval and Triplet Ranking} \label{sec: ranking}

After identifying the corresponding entities \( E_{Q'} \), we retrieve their one-hop relations from the KG (denoted as \textit{UMLS}):
\[
E_{Q'} = \{ e_i' \in V \mid \exists e_i \in E_{Q}, e_i \mapsto e_i' \}.
\]
  
Within UMLS, there exists extensive relational information, where one entity may be associated with thousands of one-hop relations. Consequently, to facilitate the extraction of the most relevant, we propose ranking methods. We encode the question \( Q \) and each triplet \( (e_i', r, e_j') \) into \( \mathbf{q}, \mathbf{r}_{ij} \) through UmlsBERT \cite{michalopoulos2021umlsbert}. Then, we explore three techniques for ranking the triplets: \\

\noindent\textbf{Similarity Ranking} We compute the similarity score between the question embedding \( \textbf{q} \) and each relation embedding \( \textbf{r}_{ij} \).

\noindent\textbf{Answer Expansion Ranking} 
We first utilize LLMs to generate a hallucinatory answer \( A \) for the question \( Q \) , and then we encode the concatenation of $[Q,A]$ to obtain text embedding \( \textbf{t} \). Subsequently, we utilize the expanded question embedding \( \textbf{t} \) to search for the most similar triplets in vector space. The detailed prompt for answer expansion can be found in Appendix~\ref{AE_prompt}.

\noindent\textbf{MMR Ranking}
This method is inspired by an information extraction method Maximal Marginal Relevance (MMR) \cite{Carbonell1998TheUO}. Initially, we identify the triplet with the highest similarity score to the question \(Q\). For the remaining triplets, we dynamically adjust their similarity scores based on the ones that have already been selected. In this way, we could consider both relevancy and redundancy:
\[
w = {w}_{base} + \delta \cdot n,
\]
\vspace{-5mm}
\[
\text{score}_{ij} = \text{sim}(\textbf{q}, \textbf{r}_{ij}) - w \cdot \overline{\text{sim}}(\textbf{r}_{ij}, \textbf{r}_{sel}).
\]
Where, \( w \) is an adjustable weight, with a base weight and \( \delta \) as the incremental weight factor per selected triplet, \( n \) is the count of triplets that have been selected. 

\noindent\textbf{Re-ranking} After the ranking stage, we obtain an ordering of the triplets. We then employ a medical cross-encoder model, MedCPT \cite{jin2023medcpt}, to re-rank them, ensuring that the most relevant triples are chosen. The re-ranked top-$p$ triplets, combined with the task prompt, are input into LLMs for answer generation. The detailed prompt can be found in Appendix~\ref{kgrank_prompt}.

\section {Experiments}
We conduct experiments on four selected medical QA datasets, in which the answers are free-text, as shown in Tab.~\ref{tab:statistics}. LiveQA \cite{abacha2017overview} consists of health questions submitted by consumers to the National Library of Medicine. It includes a training set with 634 QA pairs and a test set comprising 104 QA pairs, which is used for evaluation. ExpertQA \cite{malaviya2023expertqa} is a high-quality long-form QA dataset with 2177 questions spanning 32 fields, along with answers verified by domain experts. Among them, 504 medical questions (Med) and 96 biology (Bio) questions were used for evaluation. MedicationQA \cite{abacha2019bridging} includes 690 drug-related consumer questions along with information retrieved from reliable websites and scientific papers. We evaluate the generated answers using ROUGE \cite{lin2004rouge}, BERTScore \cite{zhang2019bertscore}, MoverScore \cite{zhao2019moverscore} and BLEURT \cite{sellam2020bleurt}. 

\begin{table}[h]
\Large
\resizebox{0.5\textwidth}{!}{
\begin{tabular}{lcccc}
\toprule
\textbf{Dataset} & Sentence (Q) & Word (Q) & Sentence (A) & Word (A) \\
\midrule
LiveQA       & 1.15 & 14.76 & 6.96 & 141.02 \\
ExpertQA-Bio & 1.26 & 21.69 & 6.18 & 184.38 \\
ExpertQA-Med & 1.37 & 22.19 & 5.96 & 180.55 \\
MedQA        & 1.02 & 7.36  & 3.38 & 71.48  \\
\bottomrule
\end{tabular}}
\caption{Statistics on the average number of sentences and words across four medical datasets (Q: Question, A: Answer).}
\label{tab:statistics}
\vspace{-5mm}
\end{table}
 
\subsection{Results}\label{sec: results}
As shown in Tab.~\ref{tab:auto_eval}, we evaluate GPT-4 and LLaMa2-13b across the following settings: zero-shot (ZS), and three proposed ranking techniques: Similarity Ranking (Sim), Answer Expansion Ranking (AE), and Maximal Marginal Relevance Ranking (MMR). Also with the Re-ranking (RR), which is on top of the Similarity Ranking. 

\begin{table*}[t!]
\Large
\centering
\resizebox{1\textwidth}{!}{
\begin{tabular}{llcccc|cccc}
\toprule
\multirow{2}{*}{\textbf{Dataset}} & \multirow{2}{*}{\textbf{Method}} & \multicolumn{4}{c|}{\textbf{GPT-4}} & \multicolumn{4}{c}{\textbf{LLaMA2-13b}} \\
\cmidrule{3-10}
 & & ROUGE-L & BERTScore & MoverScore & BLEURT 
 & ROUGE-L & BERTScore & MoverScore & BLEURT \\
\midrule
LiveQA &ZS & 18.89 & 82.50 & 54.02 & 39.84 & 17.73 & 81.93 & 53.37 & 40.45 \\
& Sim & 19.35 & \textbf{83.01} & 54.08 & 40.47 & 18.52 & 82.78 & \textbf{53.79} & \textbf{40.59} \\
& AE & 19.24 & 82.95 & 54.04 & 40.15 & 18.45 & 82.60 & 53.70 & 39.80 \\
& MMR & 19.32 & 82.91 & 54.03 & \textbf{40.55} & 18.25 & 82.70 & 53.67 & 40.22 \\
& RR & \textbf{19.44} & 82.94 & \textbf{54.11} & 40.50 & \textbf{18.83} & \textbf{82.79} & 53.72 & 39.59 \\
\midrule
ExpertQA-Bio &ZS & 23.00 & 84.50 & 56.15 & 44.53 & 23.26 & 84.38 & 55.58 & 44.65 \\
& Sim & 25.90 & 85.72 & 56.73 & 45.10 & 24.96 & 84.91 & 55.83 & 44.35 \\
& AE & 26.78 & 85.77 & 56.79 & 45.18 & 24.84 & 84.97 & 55.72 & 43.55 \\
& MMR & 26.54 & 85.76 & 56.77 & 44.93 & 25.40 & 85.08 & 55.98 & 44.04 \\
& RR & \textbf{27.20} & \textbf{85.83} & \textbf{57.11} & \textbf{45.91} & \textbf{25.79} & \textbf{85.18} & \textbf{56.17} & \textbf{45.20} \\
\midrule
ExpertQA-Med & ZS & 25.45 & 85.11 & 56.50 & 45.98 & 24.86 & 84.89 & 55.74 & 46.32 \\
& Sim & 27.61 & 86.10 & 57.13 & 46.47 & 26.40 & 85.50 & 56.23 & 46.15 \\
& AE & 27.98 & 86.12 & 57.25 & 46.80 & 26.15 & 85.36 & 56.17 & 46.02 \\
& MMR & 27.78 & 86.22 & 57.28 & 46.84 & 26.42 & 85.57 & 56.24 & 46.41 \\
& RR & \textbf{28.08} & \textbf{86.30} & \textbf{57.32} & \textbf{47.00} & \textbf{27.49} & \textbf{85.80} & \textbf{56.58} & \textbf{46.47} \\
\midrule
MedicationQA & ZS & 14.41 & 82.55 & 52.62 & 37.41 & 13.30 & 81.81 & 51.96 & 38.30 \\
& Sim & 16.05 & 83.56 & 53.23 & 37.60 & 14.60 & 82.73 & 52.47 & 38.38 \\
& AE & 16.13 & 83.46 & 53.23 & 37.87 & 14.19 & 82.50 & 52.33 & 37.90 \\
& MMR & 15.89 & 83.48 & 53.22 & 37.73 & 14.56 & 82.69 & 52.44 & 38.31 \\
& RR & \textbf{16.19} & \textbf{83.59} & \textbf{53.30} & \textbf{37.91} & \textbf{14.71} & \textbf{82.79} & \textbf{52.59} & \textbf{38.42} \\
\bottomrule
\end{tabular}}
\caption{Automatic evaluation scores: we compare ROUGE-L, BERTScore, MoverScore, BLEURT on different settings. The superior scores among the same models are highlighted in \textbf{bold}.}
\label{tab:auto_eval}
\vspace{-5mm}
\end{table*}
\subsection{Datasets}\label{sec:data}

The results show that incorporating the knowledge graph and ranking techniques notably enhances performance in almost all benchmarks and evaluation metrics in the zero-shot setting, demonstrating the effectiveness of KG-Rank. Significantly, the RR method excels in the ExpertQA-Bio, ExpertQA-Med, and Medication QA datasets, particularly evident in the over 18\% increase in the ROUGE-L score for ExpertQA-Bio. While KG-Rank still shows effectiveness on LiveQA, the RR method does not show steady improvement compared to other ranking techniques. This inconsistency may arise since the answers in LiveQA are generated via automatic extraction methods, leading to issues with semantic coherence and disorganized formats. Moreover, the performance of the three ranking methodologies exhibited variability across various datasets, indicating their unique strengths and limitations in differing contexts.

In assessing model performance, GPT-4 consistently surpasses LLaMa2-13b in both zero-shot and various ranking settings. Additionally, we evaluate the zero-shot performance of a medical LLM on these datasets in Section~\ref{med_llm} (Medical LLM). 

\section{Ablation Study and Analysis}

\paragraph{Medical LLM}
\label{med_llm}
To further investigate the capability of the medical LLM, we compare the zero-shot performance of LLaMa2-7b and baize-healthcare \cite{xu2023baize} without KG-Rank. Baize-healthcare, which is fine-tuned on LLaMa-7b using medical data, consistently outperforms LLaMa2-7b across all four datasets, as shown in Fig.~\ref{fig:zero-shot comparison}. More comparison results can be found in Appendix~\ref{sec: zero_shot_llms}.

\begin{figure}[h]
    \centering
    \begin{tikzpicture}
    \footnotesize
    \begin{axis}[
        ybar,
        bar width=14pt, 
        width=8.5cm,
        height=4cm,
        enlarge x limits=0.2, 
        legend style={at={(0.5,-0.3)},
            anchor=north,
            legend columns=-1},
        symbolic x coords={LiveQA,Ep-Bio,Ep-Med,MedicationQA},
        xtick=data,
        ymax=88,
        ymin=80,
        nodes near coords, 
        nodes near coords align={vertical},
        nodes near coords style={font=\fontsize{7}{8}\selectfont},
        /pgf/number format/.cd,  
        fixed,                     
        fixed zerofill,            
        precision=1, 
]
    \addplot[fill=color2, draw=none, draw opacity=0] coordinates {(LiveQA,81.83) 
    (Ep-Bio,84.14)
    (Ep-Med,84.72)
    (MedicationQA,81.77)};
    \addlegendentry{LLaMa2-7b}
    \addplot[fill=color1, draw=none, draw opacity=0] coordinates {(LiveQA,83.30) 
    (Ep-Bio,85.32)
    (Ep-Med,85.73)
    (MedicationQA,83.37)};
    \addlegendentry{baize-healthcare}

    \end{axis}
    \end{tikzpicture}
    \caption{BERTScore comparison: zero-shot setting with LLaMa2-7b and Baize-Healthcare. Ep stands for ExpertQA.}
    \label{fig:zero-shot comparison}
    \vspace{-5mm}
\end{figure}

\paragraph{Re-ranking Models} 
We employ GPT-4 with similarity ranking as the final setting and compare two re-ranking models: the MedCPT cross-encoder model, trained on the extensive PubMed articles, and the Cohere (\url{https://cohere.com}) re-ranking model, designed for broader domain applications. As shown in Tab.~\ref{tab:rerank com}, MedCPT steadily outperforms the Cohere re-rank model on all datasets, highlighting the importance of specialized re-rank models in the medical field. Additional evaluations are provided in Appendix~\ref{Cohere}.

\begin{table}[h]
\Large
\resizebox{0.5\textwidth}{!}{
\begin{tabular}{lcccc}
\toprule
\textbf{Dataset} & ROUGE-L & BERTScore & MoverScore & BLEURT \\
\midrule
\textit{\textbf{Cohere}} & & & & \\
LiveQA & 18.72 & 82.94 & 54.08 & 40.07 \\
ExpertQA-Bio & 26.08 & 85.81 & 56.93 & 45.70 \\
ExpertQA-Med & 27.59 & 86.08 & 57.14 & 46.54 \\
MedicationQA & 16.14 & 83.46 & 53.25 & 37.82 \\
\midrule
\textit{\textbf{MedCPT}} & & & & \\
LiveQA & \textbf{19.44} & \textbf{82.95} & \textbf{54.11} & \textbf{40.50} \\
ExpertQA-Bio & \textbf{27.20} & \textbf{85.83} & \textbf{57.11} & \textbf{45.91} \\
ExpertQA-Med & \textbf{28.08} & \textbf{86.30} & \textbf{57.32} & \textbf{46.84} \\
MedicationQA & \textbf{16.19} & \textbf{83.59} & \textbf{53.30} & \textbf{37.91} \\
\bottomrule
\end{tabular}}
\caption{The performance of Cohere re-rank model and \\ MedCPT
in the re-ranking stage.}
\label{tab:rerank com}
\vspace{-5mm}
\end{table}

\begin{figure}[h]
  \centering
  \includegraphics[width=0.5\textwidth]
  {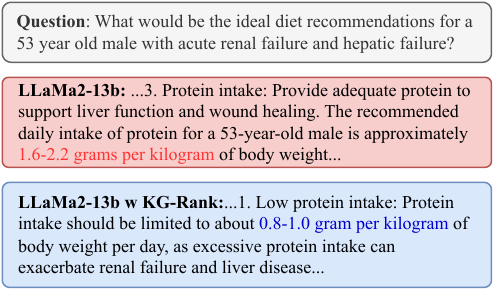} 
  \caption{A case study from ExpertQA-Med: results from LLaMa2-13b and with KG-Rank.}
  \label{fig:case_study}
  \vspace{-5mm}
\end{figure}

\paragraph{Case Study}\label{sec: case_study}
To further analyze the generated content of the KG-Rank framework, a case study is presented in Fig.~\ref{fig:case_study}. When asked about ideal diet recommendations for a 53-year-old male with acute renal failure and hepatic failure, both provide guidelines regarding protein intake. However, the original recommendation emphasizes ensuring adequate protein consumption (\textit{1.6-2.2 grams per kilogram}), whereas the answer generated under the KG-Rank framework advises controlling protein intake (\textit{limited to about 0.8-1 gram per kilogram}). The difference is critical for patients with acute renal and hepatic failure, where an inappropriate protein dosage, such as the higher range of 1.6-2.2 grams per kilogram, could worsen the strain on already compromised kidneys and liver, potentially leading to escalated health issues. This case shows that KG-Rank is more factually correct in the generated answer. More case studies can be found in the Appendix~\ref{mintaka}.

\paragraph{LLM-based Evaluation}\label{sec: gpt4 evaluation}
Although KG-Rank achieves significant improvements in ROUGE, BERTScore, MoverScore, and BLEURT, these automatic scores may have limitations in evaluating the factuality of long-form medical QA. Therefore, we introduce GPT-4 score specifically for factuality evaluation \cite{zheng2024judging}. The evaluation criteria are designed by two resident physicians with over five years of experience, which can be found in Appendix~\ref{phy_c}. As shown in Tab.~\ref{tab:gpt4-eval}, we choose GPT-4 as the vanilla model, and KG-Rank outperforms the zero-shot setting across all datasets.

\begin{table}[h]
\small
\resizebox{0.47\textwidth}{!}{
\begin{tabular}{lcccc}
\toprule
\textbf{Dataset} & Zero-Shot & Tie & KG-Rank \\
\midrule
LiveQA       & 0 &  43 & \textbf{61} \\
ExpertQA-Bio & 0 &  43 & \textbf{52} \\
ExpertQA-Med & 3 & 235 & \textbf{266} \\
MedQA        & 8 & 211 & \textbf{468}  \\
\bottomrule
\end{tabular}}
\caption{GPT-4 evaluation across four medical datasets.}
\label{tab:gpt4-eval}
\vspace{-5mm}
\end{table}

\paragraph{KG-Rank in Open Domain}\label{sec: res_open_domain}
Additionally, to demonstrate the effectiveness of our KG-Rank, we extend it to the open domain by replacing UMLS with Wikipedia through the DBpedia API (\url{https://www.dbpedia.org/}). We conduct the experiment on Mintaka \cite{sen2022mintaka}, which is a complex, natural, and multilingual dataset designed for experimenting with end-to-end question-answering models. We randomly select 1,000 pairs from the test set for evaluation. Under the enhancement of the KG-Rank framework, the accuracy increases from 60.40\% to 61.90\%. The detailed prompt can be found in Appendix~\ref{kgrank_mintaka_prompt}. 

We also conduct experiments in the domains of law, business, music, and history using the ExpertQA dataset. We employ GPT-4 as the vanilla model and use ROUGE-L, BERTScore, and MoverScore for evaluation. As shown in Tab.~\ref{tab:open_domain}, KG-Rank outperforms the baseline across all benchmarks. Building on these findings, the effectiveness of our framework is not limited to the medical domain but can also be applied to various other fields. For more case studies, please refer to Appendix \ref{mintaka}.

\begin{table}[h]
\resizebox{0.5\textwidth}{!}{
\begin{tabular}{lccccc}
\toprule
\textbf{Setting} & & & ROUGE-L & BERTScore & MoverScore \\
\midrule
\multicolumn{5}{l}{\textit{\textbf{ExpertQA-Law}}} \\
Base & & & 26.33 & 85.03 & 48.57 \\
KG-Rank & & & \textbf{29.93} & \textbf{86.25} & \textbf{48.63} \\
\midrule
\multicolumn{5}{l}{\textit{\textbf{ExpertQA-Business}}} \\
Base & & & 21.78 & 84.46 & 48.92 \\
KG-Rank & & & \textbf{24.20} & \textbf{85.42} & \textbf{49.10} \\
\midrule
\multicolumn{5}{l}{\textit{\textbf{ExpertQA-Music}}} \\
Base & & & 23.84 & 85.21 & 45.73\\
KG-Rank & & & \textbf{27.31} & \textbf{86.23} & \textbf{46.55} \\
\midrule
\multicolumn{5}{l}{\textit{\textbf{ExpertQA-History}}} \\
Base & & & 25.65 & 85.55 & 45.82 \\
KG-Rank & & & \textbf{27.75} & \textbf{86.21} & \textbf{47.08} \\
\bottomrule
\end{tabular}}
\caption{Base and KG-Rank performance in the open domain.}
\label{tab:open_domain}
\vspace{-5mm}
\end{table}

\section{Conclusion}
In this work, we propose KG-Rank, an enhanced LLM framework that integrates a medical KG and ranking techniques to improve the factuality of medical QA. As far as we know, KG-Rank is the first application of KG combined with ranking techniques for long-answer medical QA. Across four medical QA datasets, KG-Rank demonstrates over an 18\% improvement in ROUGE-L score. Its application to open domains yields a 14\% ROUGE-L score enhancement, underscoring KG-Rank's effectiveness and versatility.

\section*{Limitations}
In this research, we propose an LLM framework augmented by UMLS to improve the quality of the content generated. However, there are some limitations, which we will address in the next phase. Firstly, we plan to incorporate physician evaluations to validate the factual accuracy of KG-Rank's answers. Secondly, we aim to assess the performance of more medical-specific base models on medical QA tasks. Lastly, the ranking method may increase computational time, we recognize the need to optimize its efficiency. We will consider more graph-based methods \cite{li2022variational, yang2023going} as well as efficiency methods \cite{feng2023diffuser} later. 

\section*{Ethical Considerations}
This research utilize public medical datasets solely for academic purposes, not for practical application. We employ GPT-4, LLaMa2-13b, LLaMa2-7b, baize-healthcare for text generation, ensuring that no harmful content was produced. Both the benchmark datasets and the model outputs are free of any individual privacy data.
\bibliography{custom}

\appendix
\onecolumn
\section{Prompt Templates} \label{prompt}

In this section, we present the detailed prompt templates employed as inputs for LLMs at each phase of the KG-Rank process.

\subsection{Medical NER Prompt}\label{med_prompt}
\begin{figure}[h]
  \centering
  \includegraphics[width=1\textwidth]{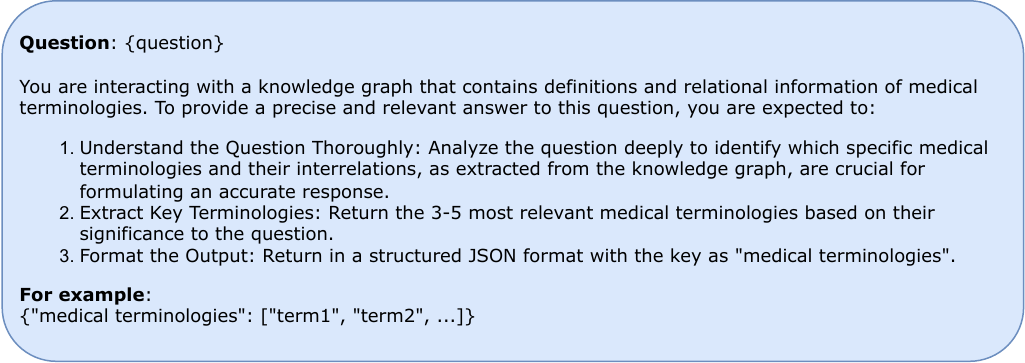} 
  \caption{Prompt used to extract medical terminologies.}
  \label{fig:ner_prompt}
\end{figure}

Fig.~\ref{fig:ner_prompt} illustrates the Medical NER prompt template that is specifically designed for extracting medical terminologies from a given question.

\subsection{Answer Expansion Prompt}\label{AE_prompt}
\begin{figure}[h]
  \centering
  \includegraphics[width=1\textwidth]{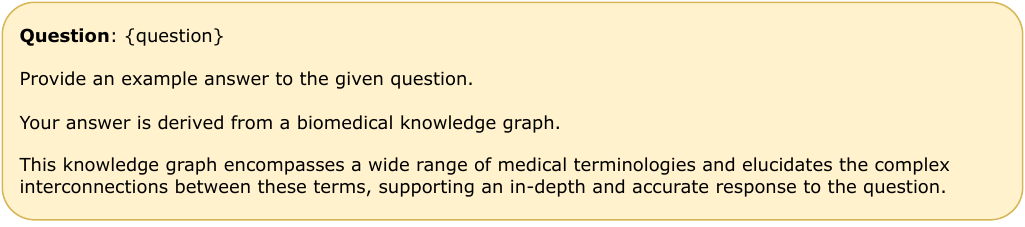} 
  \caption{Prompt for answer expansion ranking technique.}
  \label{fig:answer_expansion}
\end{figure}

Figure~\ref{fig:answer_expansion} illustrates the prompt template designed for our proposed answer expansion ranking strategy, as shown in step 2 of Fig.~\ref{fig:framework} and as described in Section~\ref{sec: ranking}.

\subsection{KG-Enhanced Prompt}\label{kgrank_prompt}
\begin{figure}[h]
  \centering
  \includegraphics[width=1\textwidth]{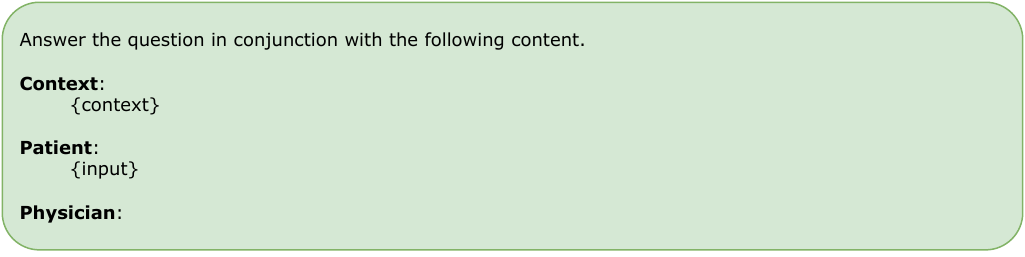} 
  \caption{Prompt for obtaining KG-enhanced LLM answers.}
  \label{fig:kg_enhanced_prompt}
\end{figure}

Fig.~\ref{fig:kg_enhanced_prompt} shows the prompt template to obtain final answers from LLMs, corresponding to step 4 in Fig.~\ref{fig:framework}.

\subsection{Physician-Designed Criteria for GPT-4 Evaluation} \label{phy_c}

Tab.~\ref{tab:phy_c} shows the criteria for evaluating medical long-form QA established by two resident physicians with over five years of experience. This critria is part of the GPT-4 evaluation prompt.

\begin{table}[h]
\centering
\begin{tabular}{p{12cm}}
\toprule
\textbf{Evaluation Criteria} \\
\midrule
\textbf{Factuality:} \\The degree to which the generated text aligns with established medical facts, providing accurate explanations for further verification. \\
\midrule
\textbf{Readability:} \\The extent to which the generated text is readily comprehensible to the user, incorporating suitable language and structure to facilitate accessibility. \\
\midrule
\textbf{Relevance:} \\The extent to which the generated text directly addresses medical questions while encompassing a comprehensive range of pertinent information. \\
\midrule
\textbf{Completeness:} \\The degree to which the generated text comprehensively portrays the clinical scenario or posed question, including other pertinent considerations. \\
\bottomrule
\end{tabular}

\caption{Physician-designed criteria for GPT-4 evaluation.}
\label{tab:phy_c}
\vspace{-3mm}
\end{table}

\subsection{KG-Enhanced Prompt for Mintaka Task}\label{kgrank_mintaka_prompt}
\begin{figure}[h]
  \centering
  \includegraphics[width=1\textwidth]{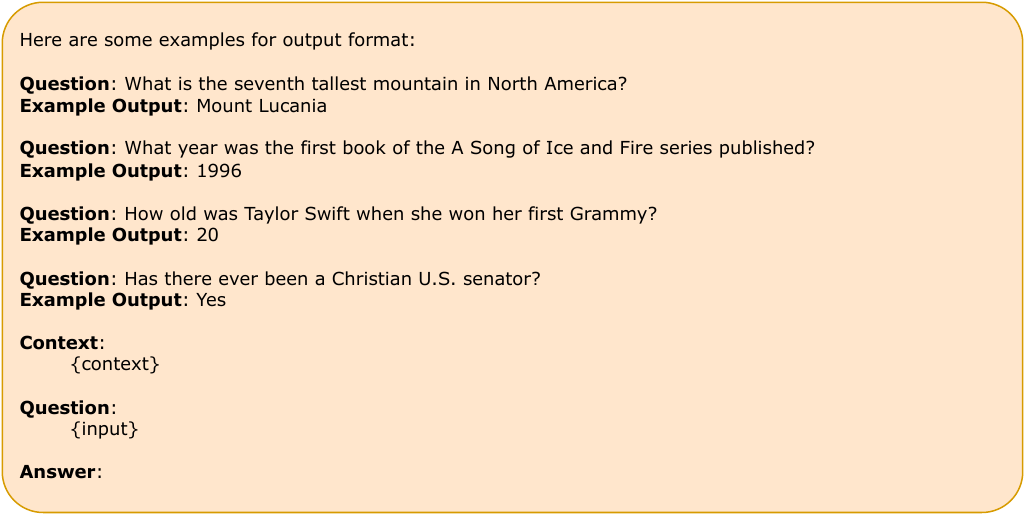} 
  \caption{Prompt for obtaining KG-enhanced LLM answers, with special design for Mintaka dataset.}
  \label{fig:KG_enhanced_prompt_for_mintaka}
\end{figure}

Fig.~\ref{fig:KG_enhanced_prompt_for_mintaka} presents the prompt for obtaining KG-enhanced LLM answers, specially designed for the Mintaka dataset.

\clearpage
\section{Detailed Evaluation Results}
\subsection{Zero-shot Performance of Different LLMs}
\label{sec: zero_shot_llms}

In this section, we evaluate the performance of widely-used LLMs on four medical datasets under the zero-shot setting. As shown in Tab.~\ref{tab:zero_shot_appendix}, the results indicate that GPT-4 performing better than the other LLMs. 


\begin{table*}[h]
\Large
\centering
\resizebox{0.8\textwidth}{!}{
\begin{tabular}{lcccccc}
\toprule
\multirow{2}{*}{\textbf{Dataset}} & \multicolumn{6}{c}{\textbf{Evaluation Metrics}} \\
\cmidrule{2-7}
& ROUGE-1 & ROUGE-2 & ROUGE-L & BERTScore & MoverScore & BLEURT \\
\midrule
\multicolumn{7}{l}{\textit{\textbf{LLaMa2-7b}}} \\
LiveQA & 18.87	& 3.60	& 17.44	& 81.83	& 53.28	& 39.43 \\
ExpertQA-Bio & 24.19  & 6.96 & 22.15	& 84.14	& 55.18	& 43.81 \\
ExpertQA-Med & 26.24 & 8.11	& 23.86	& 84.72	& 55.51	& 45.75 \\
MedicationQA & 14.19 & 2.60	& 13.12	& 81.77	& 51.94	& 37.32 \\
\midrule
\multicolumn{7}{l}{\textit{\textbf{baize-healthcare}}} \\
LiveQA & 17.92	&2.73	&16.10	&\textbf{83.30}	&53.41	&31.30 \\
ExpertQA-Bio & 23.45	&6.52	&21.31	&\textbf{85.32}	&54.95	&33.80 \\
ExpertQA-Med & 24.95	&7.21	&22.41	&\textbf{85.73}	&55.12	&34.52 \\
MedicationQA & 15.05	&2.48	&13.59	&\textbf{83.37}	&52.41	&31.39 \\
\midrule
\multicolumn{7}{l}{\textit{\textbf{LLaMa2-13b}}} \\
LiveQA & 19.15	&3.60	&17.73	&81.93	&53.37	&\textbf{40.45} \\
ExpertQA-Bio & \textbf{25.33}	&\textbf{7.92}	&\textbf{23.26}	&84.38	&55.58	&\textbf{44.65} \\
ExpertQA-Med & 27.41	&8.86	&24.86	&84.89	&55.74	&\textbf{46.32} \\
MedicationQA & 14.42	&2.62	&13.30	&81.81	&51.96	&\textbf{38.30} \\
\midrule
\multicolumn{7}{l}{\textit{\textbf{GPT-4}}} \\
LiveQA & \textbf{20.54}	& \textbf{4.65}	&\textbf{18.89}	&82.50	&\textbf{54.02}	&39.84 \\
ExpertQA-Bio & 25.06	&7.84	&23.00	&84.50	&\textbf{56.15}	&44.53 \\
ExpertQA-Med & \textbf{27.78}	&\textbf{9.49}	&\textbf{25.45}	&85.11	&\textbf{56.50}	&45.98 \\
MedicationQA & \textbf{15.52}	&\textbf{3.51}	&\textbf{14.41}	&82.55	&\textbf{52.62}	&37.41 \\
\bottomrule
\end{tabular}}

\caption{Automatic evaluation scores: we compare ROUGE-1, ROUGE-2, ROUGE-L, BERTScore, MoverScore, BLEURT on the zero-shot setting for different LLMs with medical QA tasks. The best scores are highlighted in \textbf{bold}.}
\label{tab:zero_shot_appendix}
\vspace{-3mm}
\end{table*}
\subsection{Performance of Different Re-rank Models}
\label{Cohere}

In this section, we evaluate the performance of MedCPT and the Cohere re-rank model on four medical datasets within the GPT-4 with similarity ranking setting. As shown in Table~\ref{tab:reank comparison}, the results indicate that MedCPT outperforms the Cohere re-rank model.

\begin{table*}[h]
\Large
\centering
\resizebox{0.8\textwidth}{!}{
\begin{tabular}{lcccccc}
\toprule
\multirow{2}{*}{\textbf{Dataset}} & \multicolumn{6}{c}{\textbf{GPT-4}} \\
\cmidrule{2-7}
& ROUGE-1 & ROUGE-2 & ROUGE-L & BERTScore & MoverScore & BLEURT \\
\midrule
\multicolumn{7}{l}{\textit{\textbf{Cohere}}} \\
LiveQA & 21.08 & 4.13 & 18.72 & 82.94 & 54.08 & 40.07 \\
ExpertQA-Bio & 29.07 & 9.35 & 26.08 & 85.81 & 56.93 & 45.70 \\
ExpertQA-Med & 30.84 & 10.62 & 27.59 & 86.08 & 57.14 & 46.54 \\
MedicationQA & 17.76 & 3.65 & 16.14 & 83.46 & 53.25 & 37.82 \\
\midrule
\multicolumn{7}{l}{\textit{\textbf{MedCPT}}} \\
LiveQA & \textbf{21.70} & \textbf{4.33} & \textbf{19.44} & \textbf{82.95} & \textbf{54.11} & \textbf{40.50} \\
ExpertQA-Bio & \textbf{30.05} & \textbf{10.51} & \textbf{27.20} & \textbf{85.83} & \textbf{57.11} & \textbf{45.91} \\
ExpertQA-Med & \textbf{31.34} & \textbf{10.96} & \textbf{28.08} & \textbf{86.30} & \textbf{57.32} & \textbf{46.84} \\
MedicationQA & \textbf{17.94} & \textbf{3.72} & \textbf{16.19} & \textbf{83.59} & \textbf{53.30} & \textbf{37.91} \\
\bottomrule
\vspace{-3mm}
\end{tabular}}
\caption{Automatic evaluation scores: we compare the performance of different re-rank models on ROUGE-1, ROUGE-2, ROUGE-L, BERTScore, MoverScore, BLEURT. The best scores are highlighted in \textbf{bold}.}
\label{tab:reank comparison}
\end{table*}

\clearpage
\section{More Case Studies}
\label{mintaka}
\begin{figure}[h]
  \centering
  \includegraphics[width=1\textwidth]{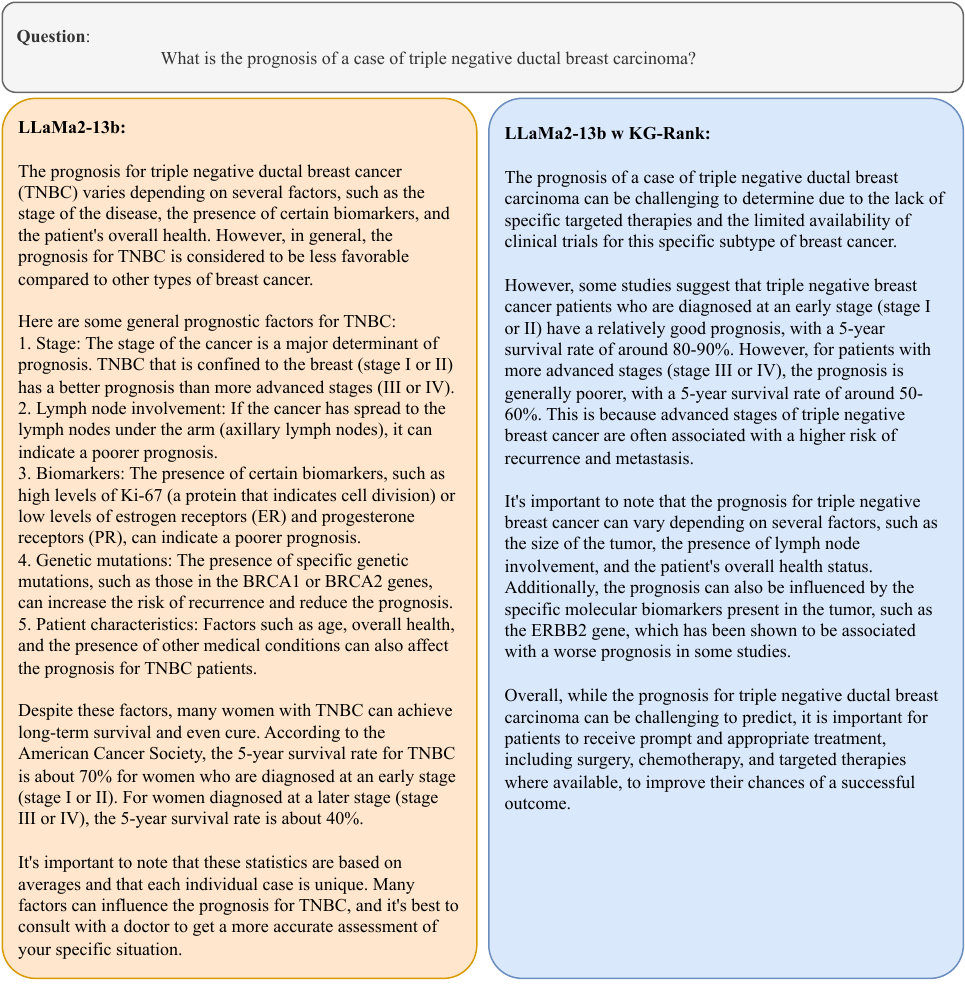} 
  \caption{A case study from ExpertQA-Med: we show results from vanilla LLaMa2-13b and KG-Rank-enhanced LLaMa2-13b.}
  \label{fig:app_med_case}
\end{figure}

\begin{figure}[h]
  \centering
  \includegraphics[width=0.5\textwidth]{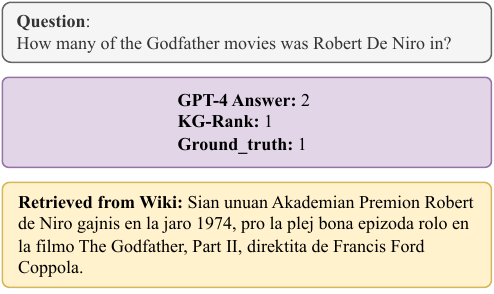} 
  \caption{A case study from Mintaka: we show results from vanilla GPT-4 and KG-Rank-enhanced GPT-4.}
  \label{fig:case_study_mintaka}
\end{figure}

We put another case study from the ExpertQA-Med dataset,  where in regards to the prognosis survival rates of breast cancer cases, the answer generated by KG-Rank is more factually accurate in terms of medical evidence, as shown in Fig.~\ref{fig:app_med_case}. Moreover, Fig.~\ref{fig:case_study_mintaka} shows a case study on the open-domain QA tasks from the Mintaka dataset, comparing the performance of the vanilla GPT-4 model against the KG-Rank-enhanced GPT-4 model. The case study involves a question: ``How many of the Godfather movies was Robert De Niro in?'' While GPT-4 responded with ``2'', our proposed KG-Rank-enhanced GPT-4 provided the correct answer ``1'', which matches the ground truth.
We also show the evidence retrieved from DBPedia. 
This case study shows that by incorporating KG-Rank, the model is able to leverage the relevant information effectively to derive the correct answer, whereas the vanilla GPT-4 did not. This demonstrates the efficacy of KG-Rank in improving the accuracy of answers in LLMs when dealing with general domain factual questions.

\section{Experimental Setup}
In our experimental setup, we employ \texttt{UmlsBERT\footnote{\url{GanjinZero/UMLSBert_ENG}},
baize-healthcare\footnote{\url{https://huggingface.co/project-baize/baize-healthcare-lora-7B}}, llama-2-7b-chat-hf\footnote{\url{https://huggingface.co/meta-llama}}, llama-2-13b-chat-hf\footnote{\url{https://huggingface.co/meta-llama}}, MedCPT\footnote{\url{https://huggingface.co/ncbi/MedCPT-Cross-Encoder}}} from Hugging Face. For GPT-4, we use the OpenAI API with a zero-temperature setting. For the Cohere re-rank model, we employ it through its API. In the MMR Ranking setting, the default value for \( w \) is 0.1, and \( \delta \) is set to 0.01. All experiments are conducted on a cluster equipped with 4 NVIDIA A100 GPUs. The prediction for each sample takes about a few seconds. Based on the size of each dataset, it may take up to hours to finish the evaluation.

\end{document}